\documentclass[letterpaper]{article}
\usepackage{aaai}
\usepackage{times}
\usepackage{helvet}
\usepackage{courier}
\usepackage{graphicx}
\usepackage{subfigure}
\usepackage{float}
\usepackage{amsmath}
\usepackage{booktabs}
\usepackage{threeparttable}
\usepackage{multirow}
\usepackage{algorithm, algorithmic}
\title{Trainable Activation Function in Image Classification}
\author{Zhaohe Liao\\
Beijing Institue of Technology\\
1120172811@bit.edu.cn\\
}
\begin{document}
\maketitle
\begin{abstract}
\begin{quote}
	In the current research of neural networks, the activation function is manually specified by human and not able to change themselves during training. This paper focus on how to make the activation function trainable for deep neural networks. We use series and linear combination of different activation functions make activation functions continuously variable. Also, we test the performance of CNNs with Fourier series simulated activation(Fourier-CNN) and CNNs with linear combined activation function (LC-CNN) on Cifar-10 dataset. The result shows our trainable activation function reveals better performance than the most used ReLU activation function. Finally, we improves the performance of Fourier-CNN with Autoencoder, and test the performance of PSO algorithm in optimizing the parameters of networks.
\end{quote}
\end{abstract}

\section[intro]{Introduction} \label{sec:Intro}
In the past researches of neural network, the core ideal is to simulate the structure of neurons in human brains. Currently, the most used neuron models, M-P neuron model \cite{Mcculloch1943A}, abstracts a neuron as a combination followed by an activation function. In the most famous models, such as ResNet \cite{Krizhevsky2012ImageNet} and R-CNN \cite{DBLP:journals/corr/GirshickDDM13}, the activation function is fixed, training approach represents the change of combination function. In the biological explanation of the M-P neuron model, the combination function represents accepting and gathering the information received from other neurons, and the activation function represents the process of passing the gathered information. However, even we have designed many kinds of complex combination function(i.e. convolution) and train its parameters, the activation function is usually simple and invariant. Can these simple activation functions simulate the complex information transform process in human brain? Can the network choose the best activation function through given data? I believe the activation function should be inferred from data, but not decided by human's prior knowledge: the features are often too complex for human to realize how different activation functions effects the result of feature extraction and partition.

However, it's a hard work to decide the time and method of changing one activation function to another (e.g. from ReLU to Sigmoid). The core question is to find when and which does one activation function performs better against another. To artificially decide these threshold means understanding the effect of each activation function on the performance of neural networks. Unfortunately, the neural network is commonly unexplainable. Thus, decide how to switch between activation functions seems to be not easy. 

Fortunately, it has been proved mathematically that a cluster of functions may be represented using a unified format: series. They are expressive and parameterized, so they can be easily converted into different functions through changing the parameters. If we could represent the activation function with series, it would be easy to adjust the activation function by choosing different parameters, which can be performed with existing optimization method(i.e. SGD, RMSProp).
In another way to implement the mixture of different activation function is combine them linearly. The final activation function is the weighted mean of several activation functions. The weights of each activation function can be trained during the given data like training weight matrices in combination function.

To try out this possibility, the experiment in Section \ref{sec:Expr} is designed. We use different activation functions, including those represented by series, in CNNs and compare their performance in image classification task on Cifar10. The result shows that the activation functions represented by series make CNNs more expressive and more generalizable than common activation functions.

To improve the performance of neural networks, we use unsupervised learning method in the pre-training of out neural networks. Autoencoder is introduced in our model to make networks well pre-trained. It is shown that autoencoder makes the training of neural networks better and faster.

Finally, we use PSO algorithm to train the network, and compare its result of training with that of BP. The result shows that PSO algorithm in small region cannot performs better then BP training methods,

\section[Expr]{Experimental Method} \label{sec:Expr}
\subsection[TaskDef]{Task Definition}
The task is quite typical: image classification on Cifar-10 dataset.

The CIFAR-10 dataset consists of 60000 32x32 color images in 10 classes, with 6000 images per class. There are 50000 training images and 10000 test images.
The dataset is divided into five training batches and one test batch, each with 10000 images. The test batch contains exactly 1000 randomly-selected images from each class. The training batches contain the remaining images in random order, but some training batches may contain more images from one class than another. Between them, the training batches contain exactly 5000 images from each class.

The performance of different solutions is mainly evaluated by the accuracy of image classification on the training batch. What's more, the speed of converge and accuracy on training batches is also considered.

\subsection[Method]{Method Description}
We solve the image classification task based on simple CNNs. 
The CNNs in this experiment has 3 sizes: small, middle, and large. They all have the structure shown in Figure 1 below.

\begin{figure}[H]
	\centering
	\includegraphics[scale=0.10]{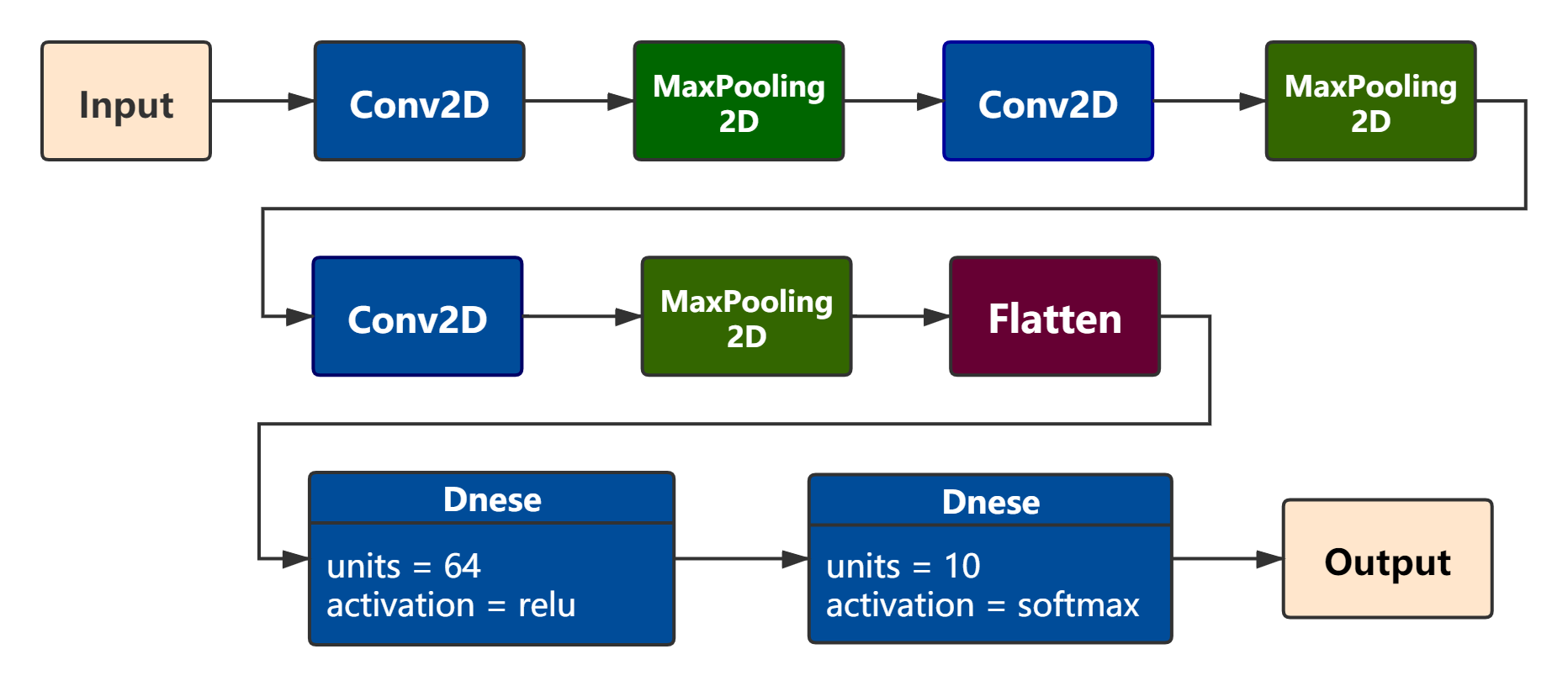}
	\caption{Structure of CNNs}
\end{figure}

The different of such CNNs are their filter numbers of convolution layers. Those numbers for convolution layers in small net are 16, 32 and 32. Those for middle net are 32, 64 and 64, and in large net, they are 48, 96 and 96.

However, as described in Section \ref{sec:Intro}, we believe CNNs with simple activation functions are not powerful enough.
To compare the effect of different activation functions, each size of CNN derived 3 CNNs with different activation function: ReLU, Fourier series, and linear combination of different simple activation function.
Thus, we have totally 9 CNNs to compare.

The CNNs are all trained with RMSProp algorithm, in which $\rho = 0.95$ and learning rate is: 
\begin{equation}
lr =\left\{
\begin{array}{rcl}
0.001 & & {0 < epoch \leq 20 }\\
0.0001 & & {20 < epoch \leq 40 }\\
0.00001 & & {40 < epoch \leq 60 }\\
0.000001 & & {60 < epoch}
\end{array} \right.
\end{equation}

The loss function is defined as the cross entropy between probabilities given by the network and one-hot labels.
The 5 Cifar-10 train datasets are merged into one dataset ,shuffled, batched, and then directly used in training, without data augmentation. 

\subsubsection[Fourier]{Fourier Series Simulated Activation Function}
In Section \ref{sec:Intro}, we propose that series can be used to construct trainable activation function. We call CNN using Fourier series as its activation function \textit{Fourier-CNN}. We choose Fourier series for it's good convergence: activation function naturally satisfy the Dirichlet Fourier series conditions, which means that Fourier series can represent any suitable activation function. We simply prove such consequence as below.

\noindent Dirichlet Fourier series conditions says: A function that

1. is absolutely integrable over a period.

2. is of bounded variation in any given bounded interval.

3. has a finite number of discontinuities that cannot be infinite in any given bounded interval.

\noindent can be expanded in a Fourier series which converges to the function at continuous points and the mean of the positive and negative limits at points of discontinuity.  

For a activation function, since it must be continuous, or noncontinuous points will make the output of neuron meaningless, it is absolutely integrable, and has zero discontinuities. Also, activation function should have finite extreme point, as for the neuron should be stable for similar input. For this reason, it is of bounded variation in any given bounded interval. Therefor, activation functions satisfy the Dirichlet Fourier series conditions. Thus, Fourier series can represent any function that is suitable for activation function, including those we often use(e.g. ReLU, Sigmoid and tanh). That is to say, theoretically, the best performance of network with Fourier series as activation function will never be worse than any activation function. 

The Fourier series can be written as
\begin{equation}
act(x) = A + \sum_{n = 1}^{\infty}(a_n\cos(n\omega x) + b_n\sin(n\omega x))
\end{equation}
\noindent and the parameters $A, \omega, a_n, b_n$ are trainable parameters. Usually, the rank of Fourier series doesn't need to be so high. In our experiment, it is fixed to 5 (i.e. $n = 1,2,...,5$).

The training approach of such activation function is simple too. It can be trained using gradient descent algorithms. The gradient of such activation function is 

\begin{equation}
\frac{\partial act(x)}{\partial A} = 1
\end{equation}

\begin{equation}
\frac{\partial act(x)}{\partial a_n} = -n\omega\sin(n\omega x)
\end{equation}

\begin{equation}
\frac{\partial act(x)}{\partial b_n} = n\omega\cos(n\omega x)
\end{equation}

\begin{equation}
\begin{split}
\frac{\partial act(x)}{\partial \omega} & = \sum_{n = 1}^{\infty}(a_n n  x\cos(n\omega x) - b_n n x\sin(n\omega x) \\
& =\sum_{n = 1}^{\infty}n x(a_n \cos(n\omega x) - b_n \sin(n\omega x))
\end{split}
\end{equation}

That is to say, if we are able to use a little memory to store $\sin(n\omega x), \cos(n\omega x)$ and $x$, the task of gradient computing sill be greatly simplified and become just computing several multiplies. The memory and time complexity during training are both $O(n)$, and $n$ is usually a very small integer representing the rank of Fourier series.

\subsubsection[Linear]{Linear Combination of Activation Functions}
Another way to make the activation function trainable is to linearly combine the multiple functions, which are called \textit{candidate functions}. CNNs using such activation are called \textit{LC-CNN}. Mathematically, such activation function is to compute the dot product of a unit vector in hyperspace and the vector of different activation functions. That is,

\begin{equation}
\begin{split}
act(x) & = \frac{\sum_{i=1}^{n} w_i act_i(x)}{\sum_{i=1}^{n} w_i} \\
& = \begin{bmatrix} w_1 & w_2 & ... & w_{n - 1} & w_n \end{bmatrix} \times \\
& \begin{bmatrix} act_1(x) & act_2(x) & ... & act_{n - 1}(x) & act_n(x) \end{bmatrix}^T \\
& = \frac{W \times Acts}{\begin{Vmatrix} W \end{Vmatrix}} 
\end{split}
\end{equation}
, where $\begin{Vmatrix} W \end{Vmatrix} = \sum_{i=1}^{n} w_i$.

There are two reasons of choosing the combination to be linear. One is that such combination can easily turns into one of the simple activation functions by just letting the weight vector $W$ to be one-hot. This means it's easy to guarantee that such method will not perform worse than any single activation function. Another is the gradient of these functions are easy to compute. In the example above, the gradients are:

\begin{equation}
\begin{split}
\frac{\partial act(x)}{\partial w_k} & = \frac{act_k(x)\sum_{i=1}^{n} w_i - \sum_{i=1}^{n} w_i act_i(x)}{(\sum_{i=1}^{n} w_i)^2} \\
& = \frac{act_k(x)\sum_{i=1}^{n} w_i - act(x) \sum_{i=1}^{n} w_i}{(\sum_{i=1}^{n} w_i)^2} \\
& = \frac{act_k(x) - act(x)}{\begin{Vmatrix} W \end{Vmatrix}}
\end{split}
\end{equation}

So if we store the sum of weights and output of each activation function, it's a easy task to compute the gradient. Thus, the efficiency of gradient decent training algorithms will not be too bad.

In our experiment, the activation function is the combination of ReLU, Sigmoid, tanh and linear.

\section[Result]{Result} \label{sec:Result}
After training on Cifar-10 dataset, different methods shows different performance. The best accuracy and loss of each networks is shown in the Table \ref{tab:performance_comparison}.

\renewcommand{\arraystretch}{1.5} 
\begin{table}[tp]
	\centering
	\fontsize{6.5}{8}\selectfont
	\begin{threeparttable}
		\caption{Best performance comparison by different size and activation function.}
		\label{tab:performance_comparison}
		\begin{tabular}{lcccc}
			\toprule
			\multirow {2}{*}{Method}&
			\multicolumn{2}{c}{Training}&\multicolumn{2}{c}{Validation}\cr
			\cmidrule(lr){2-3} \cmidrule(lr){4-5}
			&Accuracy&Loss&Accuracy&Loss\cr
			\midrule
			Small CNN&0.6789&2.752&0.5914&1.918\cr
			Small Fourier-CNN&0.7077&2.136&0.6423&1.820\cr
			Small LC-CNN&0.6910&2.238&0.6337&1.958\cr
			Middle CNN&0.8355&2.665&0.6368&1.755\cr
			Middle Fourier-CNN&0.7866&2.302&0.6738&2.274\cr
			Middle LC-CNN&0.8231&2.082&0.6532&1.959\cr			
			Large CNN&0.9578&3.501&0.6474&1.755\cr
			Large Fourier-CNN&0.9421&2.038&0.6601&2.274\cr
			Large LC-CNN&0.9583&2.302&0.6791&1.833\cr
			\bottomrule
		\end{tabular}
	\end{threeparttable}
\end{table}


Figure 2 to figure 7 shows the accuracy and loss curve of different groups of CNNs while training and predicting.

\begin{figure}[htbp]
	\centering
	
	\subfigure[accuracy]{
		\begin{minipage}[t]{0.5\linewidth}
			\centering
			\includegraphics[scale=0.3]{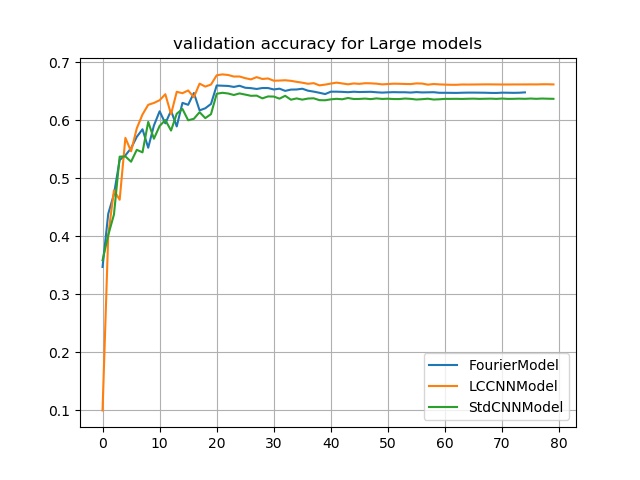}
		\end{minipage}%
	}%
	\subfigure[loss]{
	\begin{minipage}[t]{0.5\linewidth}
		\centering
		\includegraphics[scale=0.3]{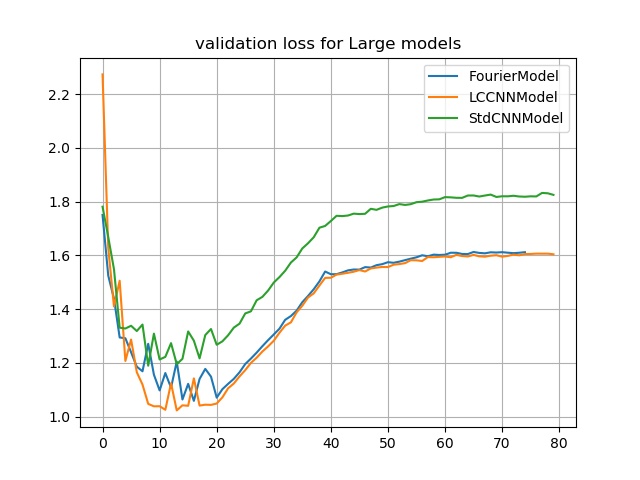}
	\end{minipage}%
	}%
	
	\centering
	\caption{Performance of large models on validation data}
\end{figure}

\begin{figure}[htbp]
	\centering
	
	\subfigure[accuracy]{
		\begin{minipage}[t]{0.5\linewidth}
			\centering
			\includegraphics[scale=0.3]{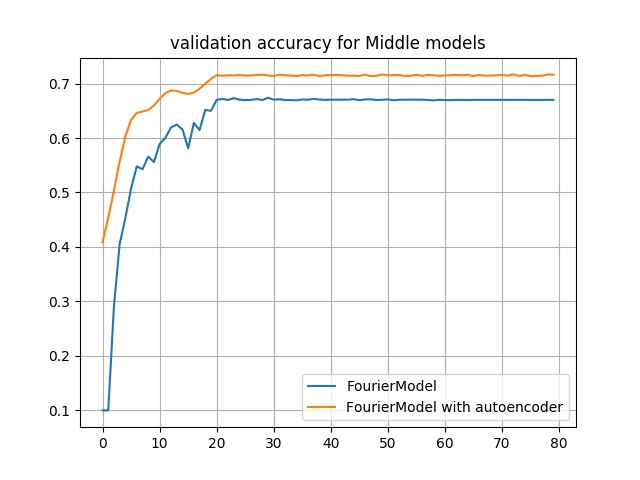}
		\end{minipage}%
	}%
	\subfigure[loss]{
		\begin{minipage}[t]{0.5\linewidth}
			\centering
			\includegraphics[scale=0.3]{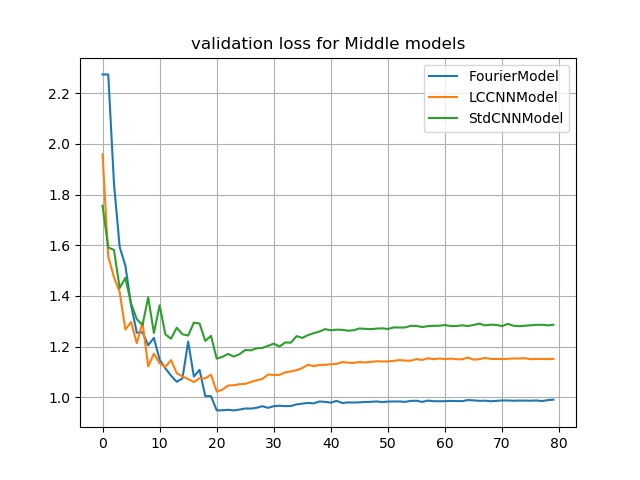}
		\end{minipage}%
	}%
	
	\centering
	\caption{Performance of middle models on validation data}
\end{figure}

\begin{figure}[htbp]
	\centering
	
	\subfigure[accuracy]{
		\begin{minipage}[t]{0.5\linewidth}
			\centering
			\includegraphics[scale=0.3]{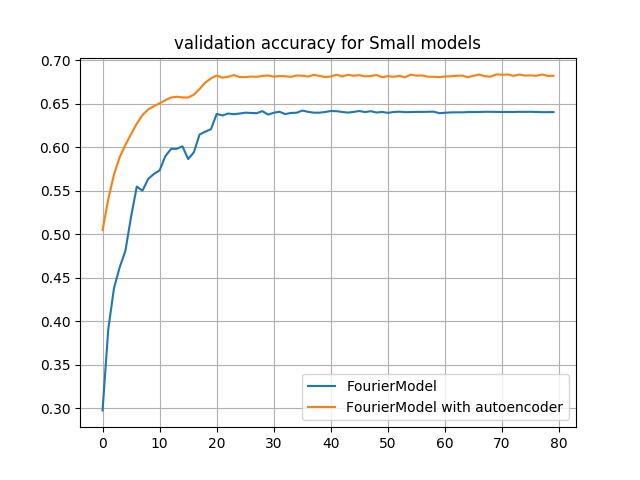}
		\end{minipage}%
	}%
	\subfigure[loss]{
		\begin{minipage}[t]{0.5\linewidth}
			\centering
			\includegraphics[scale=0.3]{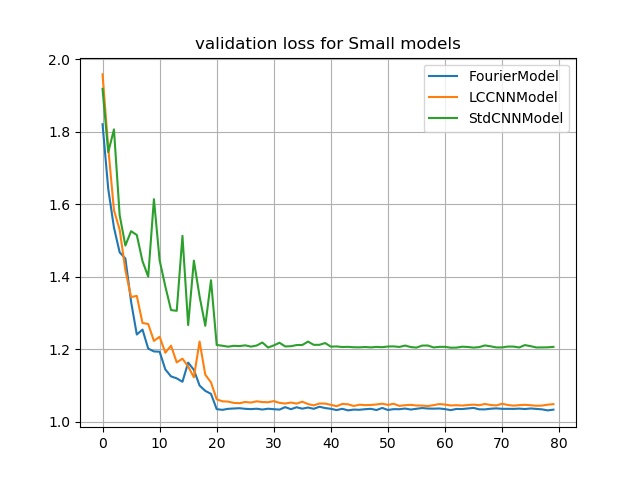}
		\end{minipage}%
	}%
	
	\centering
	\caption{Performance of small models on validation data}
\end{figure}

\begin{figure}[htbp] 
	\centering
	
	\subfigure[accuracy]{
		\begin{minipage}[t]{0.5\linewidth}
			\centering
			\includegraphics[scale=0.3]{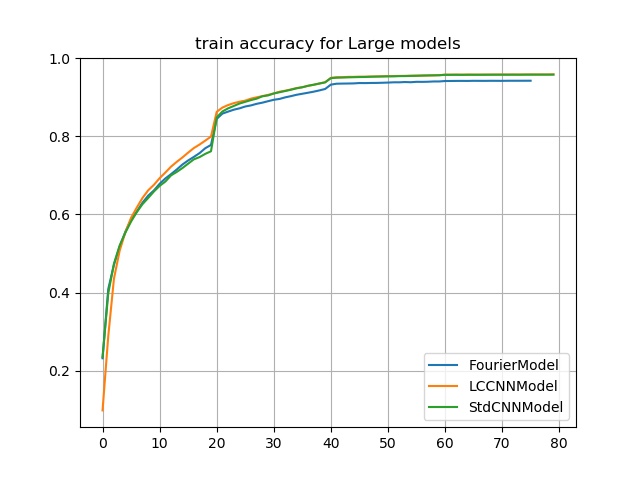}
		\end{minipage}%
	}%
	\subfigure[loss]{
		\begin{minipage}[t]{0.5\linewidth}
			\centering
			\includegraphics[scale=0.3]{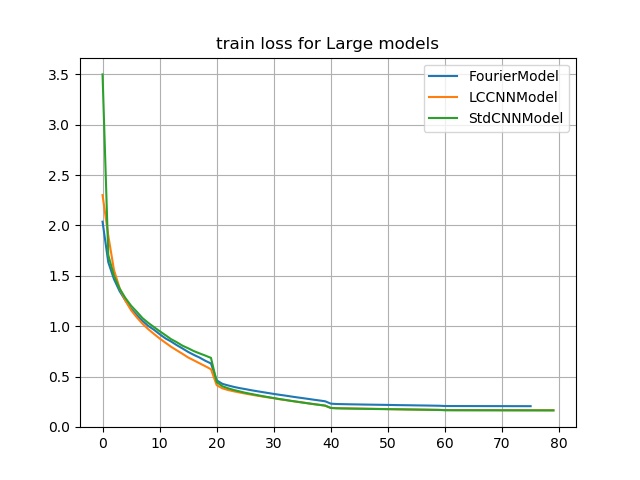}
		\end{minipage}%
	}%
	
	\centering
	\caption{Performance of large models on train data}
	\label{fig:Performance_Large_train}
\end{figure}

\begin{figure}[htbp]
	\centering
	
	\subfigure[accuracy]{
		\begin{minipage}[t]{0.5\linewidth}
			\centering
			\includegraphics[scale=0.3]{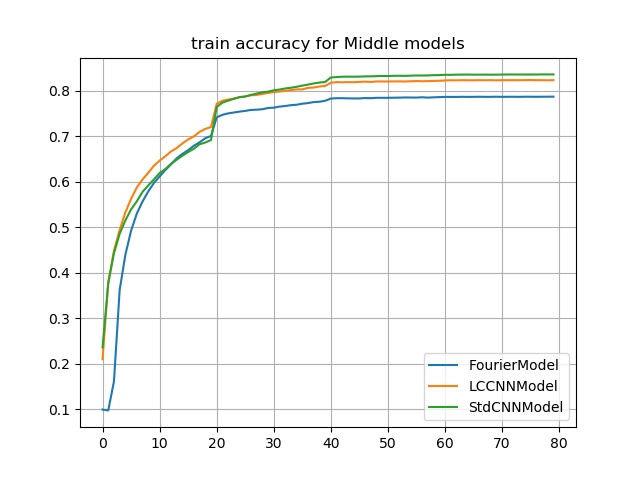}
		\end{minipage}%
	}%
	\subfigure[loss]{
		\begin{minipage}[t]{0.5\linewidth}
			\centering
			\includegraphics[scale=0.3]{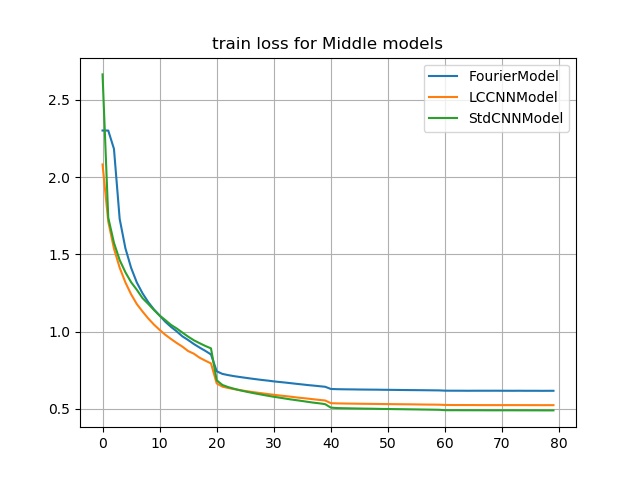}
		\end{minipage}%
	}%
	
	\centering
	\caption{Performance of middle models on train data}
\end{figure}

\begin{figure}[htbp]
\centering

\subfigure[accuracy]{
	\begin{minipage}[t]{0.5\linewidth}
		\centering
		\includegraphics[scale=0.3]{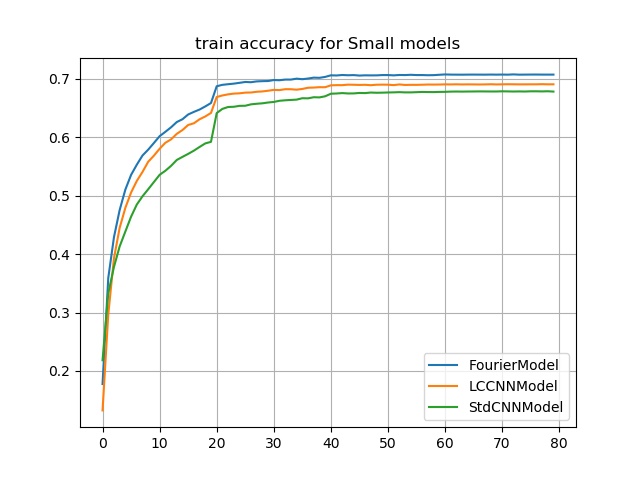}
	\end{minipage}%
}%
\subfigure[loss]{
	\begin{minipage}[t]{0.5\linewidth}
		\centering
		\includegraphics[scale=0.3]{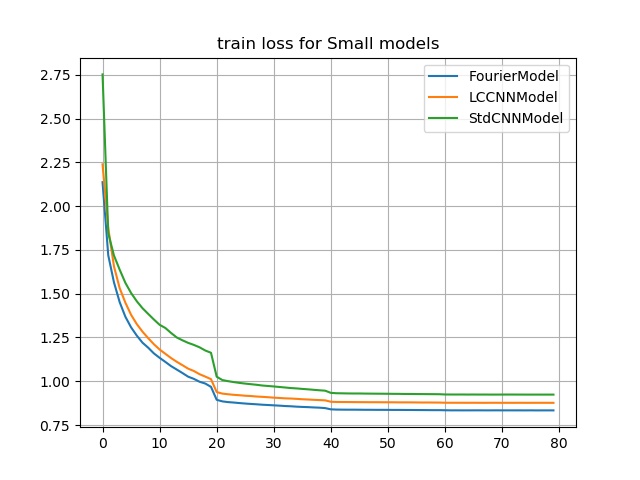}
	\end{minipage}%
}%

\centering
\caption{Performance of small models on train data}
\end{figure}

In these figures, we can see that in any size of models, the trainable activation functions gains higher accuracy and lower loss than the standard CNNs.
The improvement on accuracy of Fourier-CNN and LC-CNN compared with standard CNN in different size are concluded in Table \ref{tab:improvement}.

\renewcommand{\arraystretch}{1.5} 
\begin{table}[tp]
	\centering
	\fontsize{6.5}{8}\selectfont
	\begin{threeparttable}
		\caption{Accuracy improvement against standard CNNs in different size (absolute percentage)}
		\label{tab:improvement}
		\begin{tabular}{lcccc}
			\toprule
			\multirow {2}{*}{Size}&
			\multicolumn{2}{c}{Validation}&\multicolumn{2}{c}{Training}\cr
			\cmidrule(lr){2-3} \cmidrule(lr){4-5}
			&Fourier-CNN(\%)&LC-CNN(\%)&Fourier-CNN(\%)&LC-CNN(\%)\cr
			\midrule
			Small  &5.09&4.23&2.88&1.21\cr
			Middle &3.70&1.64&-4.89&-1.24\cr
			Large  &1.27&3.17&-1.57&0.05\cr
			\bottomrule
		\end{tabular}
	\end{threeparttable}
\end{table}

This table shows that in any size of networks, trainable activation functions reaches a higher accuracy on validation, and it's improvement is more significant on smaller networks. In small networks, Fourier-CNNs performs better than LC-CNNs and in larger networks, LC-CNNs work better.

It also shows that in larger networks, simple CNNs shows a higher accuracy on training data but lower accuracy on validation.
In the same time, Fourier-CNN reaches the best validation accuracy with the worse training accuracy between networks with middle size, and LC-CNN reaches the best accuracy on both training and validation between large networks. 

\section[Discuss]{Discussion}
\subsection[Explain]{Explaination}
We believe that it is the improvement of expressiveness and generalization ability that cause the result described in Section \ref{sec:Result}. In small models, there are only a few parameters in the model, and the models are underfitting. In this time, Fourier-CNNs and LC-CNNs improves the performance on both training and validating, which illustrate that the model's expressiveness has been improved. 

In middle models, with the improvement of the number of parameters, the expressiveness is not the bottleneck, and it is the generalization ability  that limits the performance on validation data. Thus, the factor that models with trainable activation functions reaches higher accuracy on validation data but lower accuracy on training data with almost same number of parameters with standard CNNs illustrates that Fourier-CNNs and LC-CNNs has better generalization ability.

In large models, the number of parameter becomes larger and all models reaches a high accuracy on training data. Though the accuracy on training data are almost the same (shown in Figure \ref{fig:Performance_Large_train}), the Fourier-CNN and LC-CNN reaches a better accuracy on validation data. This also proves that the trainable activation functions extends the models' generalization ability.

\subsection[Lim]{Limitation}
We are not able to explain why the difference of standard models and Fourier-CNNs becomes less and less with the improvement of model's size. Also the size of this experiment only changes the parameter in a layer, but not changes the depth of networks. Thus, the performance of trainable activation functions in deeper networks is not discussed in this paper.

In the same time, we only apply trainable activation functions on simple CNNs, thus it doesn't reaches the state-of-art performance. But this is a pervasive technique. What if use this technique on the best models we have had? Will it improves its performance? That's a question worth discussion.

\section[Improve]{improvement}
After the experiment above, we use unsupervised learning method and swarm intelligence algorithm on our models to improve it's performance.

\subsection[AutoEnc]{Autoencoder}
\begin{figure}[H]
	\centering
	\includegraphics[scale=0.52]{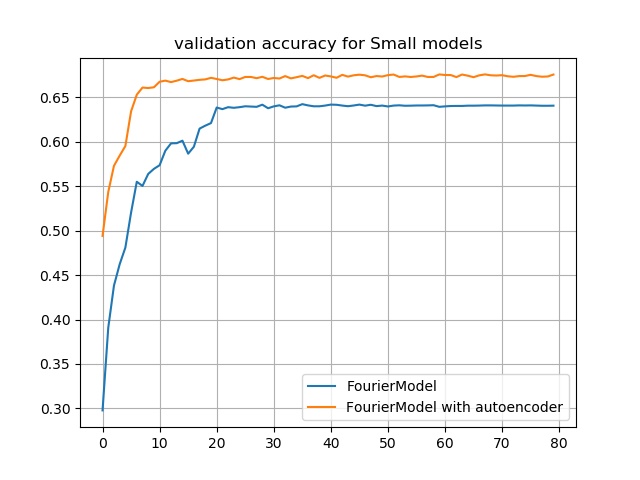}
	\caption{validation accuracy improvement with autoencoder}
	\label{fig:AutoEncoder}
\end{figure}

Autoencoder \cite{DBLP:journals/corr/DizajiHH17} is a unsupervised learning method for clustering. It forces the layer to extract the feature of original image that can be used to reconstruct the image. It evaluates the performance of layers by using the difference of image reconstructed and the original image. This technique can be used to pre-train the neural networks. In our experiment, autoencoder is used to pre-train each layer of the middle sized Fourier-CNN networks.After the pre-training of autoencoder, the result shows in Figure \ref{fig:AutoEncoder}.

Because of the pre-training, the validation accuracy starts from a higher value. The experiment shows that the training process after pre-training of autoencoder is more smooth and can reach a higher level. The reason is layers after pre-training 
has had the ability to extract the feature of images, and has been around a good minimal of loss function. Thus, networks after such pre-training can reach a better accuracy faster.

\subsection[PSO]{Particle Swarm Optimization}
In the sections above, we train the networks with gradient descent method. However, there are also many other training methods to optimize the parameters in neural networks. (e.g. evolutionary algorithms and swarm intelligence algorithms). We try to use Particle Swarm Optimization(PSO) instead of BP to optimize the parameters in standard middle-sized CNN. In PSO, each particle is a matrix that represents all parameters in the networks. Thus, a particle represents a answer to the optimize problem.
The loss of network is used to calculate the fitness $f(x)$ as 
\begin{equation}
f(x) = \frac{1}{1+\frac{1}{2n}\sum_{p=1}^{n}\left(y_p - t_p\right)}
\label{equ:fitness}
\end{equation}
, where $n$ is the number of samples and $y_p$ is the output of network, $t_p$ is the true value given by label.
The formula we use to update the parameter matrix of particle k is:
\begin{equation}
\begin{split}
V_k & = \omega V_k + C_1 rand(0,1)(P_{lk} - X_k) \\
& + C_2 rand(0,1)(V_g - X_k) \\
X_k & = X_k + V_k
\end{split}
\label{equ:pso_update}
\end{equation}
, where $\omega$ is a positive number called momentum factor, $C_1,C_2$ is constant $2$, $P_{lk}$ is the local best found by particle k, and $P_g$ is the global best found by all particles.
This algorithm works as algorithm \ref{alg:PSO}.

\begin{algorithm}
	\renewcommand{\algorithmicrequire}{\textbf{Input:}}
	\renewcommand{\algorithmicensure}{\textbf{Output:}}
	\caption{Partical Swarm Optimization Algorithm}
	\label{alg:PSO}
	\begin{algorithmic}[1]
		\REQUIRE Particles represented as tensors
		\ENSURE Best particles
		\STATE Randomly initialize all particles
		\STATE Use the tensors that represents particles as the weight of networks 
		\WHILE{in computition bound}
		\STATE Input data into networks, get the output of network $y_p$
		\STATE Calculate the fitness $f(x)$ for all particles using formula (\ref{equ:fitness})
		\STATE Update the value of each particle using the formula (\ref{equ:pso_update})
		\STATE Update $P_{lk}$, $P_g$
		\ENDWHILE
		\STATE \textbf{return} $P_g$
	\end{algorithmic}  
\end{algorithm}

\begin{figure}[H]
	\centering
	\includegraphics[scale=0.52]{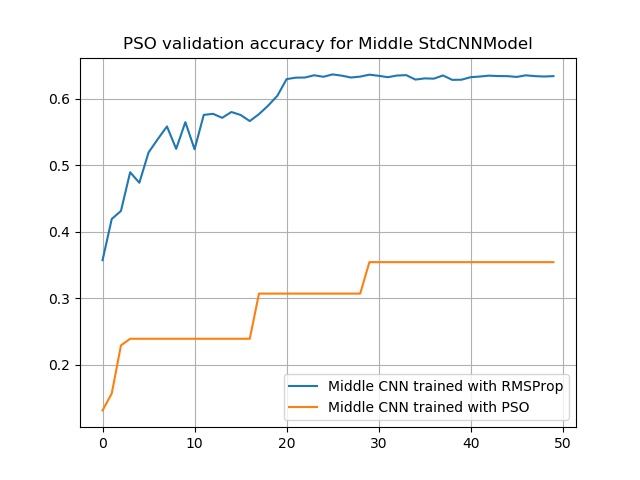}
	\caption{validation accuracy of CNN trained by PSO}
	\label{fig:PSO_vali}
\end{figure}

We set the swarm to have 10 particles, and let the computation bound to be iterating 50 generations.
The best accuracy we found (generated by $P_g$) in each iteration is shown in figure \ref{fig:PSO_vali}.

In Figure \ref{fig:PSO_vali}, we can see that the validation accuracy of CNN trained by PSO algorithm is lower than CNN trained with BP algorithm (in this case, it's RMSProp). The reason might be the size of swarm is not big enough. What's more, the update of $P_g$ is not so frequently, thus the improvement of networks is quite discrete. In most iterations, the $P_g$ doesn't updates.

\section[conclu]{Conclusion}
The long term goal of this paper is to study the way that human learns, and construct the intelligence system that simulates human's learning process. Specifically, we propose that the activation function should be trainable and give out two ways to implement it. By using Fourier series and linear combination of simple activation functions, we make the activation function variable and can be trained with BP algorithms. These methods is available for all neuron network models that uses M-P neuron models. Also, we proved that such activation functions will not be worse than simple non-trainable activation functions if it is properly trained. In specific, we test the performance of these activation functions in CNN. 

Our result shows that both ways of constructing trainable activation function are feasible and can improve the performance of CNNs in image classification task. Specifically, Fourier-CNNs shows the best performance in small networks while LC-CNNs turns to be the best on large networks. 

What's more, it's been verified in this paper that autoencoder can significantly improve the performance of networks and make them converge faster. And using PSO algorithm to train neural network seems not a good solution  as for training using PSO only reaches about half of the accuracy that BP algorithms can reach.

There are a number of future work directions that we intend to purse. For example, the method we proposed is just applied on very simple CNNs, and the data used is also not per-processed. Whether this method can improve the performance of the state-of-art models? How to ensure that Fourier series simulated activation functions will not be stacked in to a local best that is worse than simple activation functions? The answer to these questions will leads to the improvement of this method.

\bibliography{ref}

\begin{thebibliography}{}

\bibitem[\protect\citeauthoryear{Dizaji, Herandi, and
  Huang}{2017}]{DBLP:journals/corr/DizajiHH17}
Dizaji, K.~G.; Herandi, A.; and Huang, H.
\newblock 2017.
\newblock Deep clustering via joint convolutional autoencoder embedding and
  relative entropy minimization.
\newblock {\em CoRR} abs/1704.06327.

\bibitem[\protect\citeauthoryear{Girshick \bgroup et al\mbox.\egroup
  }{2013}]{DBLP:journals/corr/GirshickDDM13}
Girshick, R.~B.; Donahue, J.; Darrell, T.; and Malik, J.
\newblock 2013.
\newblock Rich feature hierarchies for accurate object detection and semantic
  segmentation.
\newblock {\em CoRR} abs/1311.2524.

\bibitem[\protect\citeauthoryear{Krizhevsky, Sutskever, and
  Hinton}{2012}]{Krizhevsky2012ImageNet}
Krizhevsky, A.; Sutskever, I.; and Hinton, G.
\newblock 2012.
\newblock Imagenet classification with deep convolutional neural networks.
\newblock {\em Advances in neural information processing systems} 25(2).

\bibitem[\protect\citeauthoryear{Mcculloch and Pitts}{1943}]{Mcculloch1943A}
Mcculloch, W.~S., and Pitts, W.
\newblock 1943.
\newblock A logical calculus of the ideas immanent in nervous activity.
\newblock {\em Bulletin of Mathematical Biology} 52(1–2):99--115.

\bibitem[\protect\citeauthoryear{Zeiler and *}{2012}]{Zeiler2012ADADELTA}
Zeiler, M.~D., and *.
\newblock 2012.
\newblock Adadelta: An adaptive learning rate method.
\newblock {\em Computer Science}.

\end{thebibliography}
\bibliographystyle{aaai}
\end{document}